%% file: main.tex
\documentclass[pdflatex,sn-basic]{sn-jnl}

\usepackage{graphicx}%
\usepackage{multirow}%
\usepackage{amsmath,amssymb,amsfonts}%
\usepackage{amsthm}%
\usepackage{mathrsfs}%
\usepackage[title]{appendix}%
\usepackage{xcolor}%
\usepackage{textcomp}%
\usepackage{manyfoot}%
\usepackage{booktabs}%
\usepackage{algorithm}%
\usepackage{algorithmicx}%
\usepackage{algpseudocode}%
\usepackage{listings}%

\usepackage{lmodern,textcomp}
\usepackage{tcolorbox}
\usepackage{tikz}
\usepackage{array}

\newcolumntype{P}[1]{>{\raggedright\arraybackslash}p{#1}}

\begin{document}

\title[Towards Adaptive Feedback with AI]{Towards Adaptive Feedback with AI: Comparing the Feedback Quality of LLMs and Teachers on Experimentation Protocols}


\author*[1]{\fnm{Kathrin} \sur{Seßler}}\email{kathrin.sessler@tum.de} 
\author[2]{\fnm{Arne} \sur{Bewersdorff}} 
\author[2]{\fnm{Claudia} \sur{Nerdel}} 
\author[1]{\fnm{Enkelejda} \sur{Kasneci}} 

\affil[1]{\orgdiv{Chair for Human-centered Technologies for Learning}}

\affil[2]{\orgdiv{Professorship Life Science Education}}

\affil{\orgname{Technical University of Munich}, \orgaddress{\city{Munich}, \country{Germany}}}



\abstract{\input{00_abstract}}




\maketitle

\input{01_introduction}
\input{02_theory}

\input{03_methods}

\input{04_results}
\input{05_discussion}
\input{06_conclusion}

\backmatter





\section*{Declarations}
The science education experts and the science teachers voluntarily participated in the study. All participants provided informed consent before participating. The study was designed to minimize any potential risks or discomfort for the participants. All procedures were in accordance with the ethical standards of the German Society of Psychology (DGPs). 
Student data was obtained from the publicly available dataset from \citet{bewersdorff2023assessing}.

\subsection{Author Contribution Information}
\textbf{Kathrin Seßler}: Writing - original draft, Writing - review \& editing, Visualization, Data curation, Methodology, Investigation, Conceptualization, Software. \textbf{Arne Bewersdorff}: Writing - original draft, Writing - review \& editing, Data curation, Methodology, Investigation, Conceptualization. \textbf{Claudia Nerdel}: Writing - review \& editing, Supervision, Conceptualization. \textbf{\mbox{Enkelejda} Kasneci}: Writing - review \& editing, Supervision, Project administration, Conceptualization.

\subsection{Declaration of competing interest}
The authors declare that they have no known competing financial interests or personal relationships that could have appeared to influence the work reported in this paper.

\subsection{Declaration of AI and AI-assisted technologies in the writing process}
During the preparation of this work, the authors used ChatGPT (GPT-4 and o1) as well as Grammarly in order to improve the readability, structure and language of sentences as some authors are not native English speakers. After using these tools, the authors reviewed and edited the content as needed and took full responsibility for the content of the publication.


\bibliography{main}

\end{document}

%% file: 00_abstract.tex
Effective feedback is essential for fostering students' success in scientific inquiry. With advancements in artificial intelligence, large language models (LLMs) offer new possibilities for delivering instant and adaptive feedback. However, this feedback often lacks the pedagogical validation provided by real-world practitioners. 
To address this limitation, our study evaluates and compares the feedback quality of LLM agents with that of human teachers and science education experts on student-written experimentation protocols.
Four blinded raters, all professionals in scientific inquiry and science education, evaluated the feedback texts generated by 1) the LLM agent, 2) the teachers and 3) the science education experts using a five-point Likert scale based on six criteria of effective feedback: Feed Up, Feed Back, Feed Forward, Constructive Tone, Linguistic Clarity, and Technical Terminology. 
Our results indicate that LLM-generated feedback shows no significant difference to that of teachers and experts in overall quality.
However, the LLM agent's performance lags in the Feed Back dimension, which involves identifying and explaining errors within the student’s work context. Qualitative analysis highlighted the LLM agent’s limitations in contextual understanding and in the clear communication of specific errors.
Our findings suggest that combining LLM-generated feedback with human expertise can enhance educational practices by leveraging the efficiency of LLMs and the nuanced understanding of educators.

%% file: 01_introduction.tex
\section{Introduction}

Effective feedback is a cornerstone of student development, influencing learning outcomes, student satisfaction and motivation \citep{gan2021teacher,wisniewski2020power,monteiro2021creating}. Personalization and contextualization significantly enhance intrinsic motivation and engagement in learning tasks \citep{cordova1996intrinsic}. However, traditional automatic feedback systems often fall short by providing generic, static responses that lack adaptiveness and fail to meet individual learner needs.

Advancements in generative Artificial Intelligence (AI), particularly large language models (LLMs), offer new opportunities to significantly improve feedback mechanisms in education \citep{kasneci2023chatgpt}. LLMs can overcome previous limitations by delivering adaptive, real-time, and personalized feedback that mimics human-like interaction. This integration combines the efficiency and scalability of computer-based feedback with the nuanced, adaptive qualities traditionally associated only with human educators. In their review, \citet{maier2022personalized} highlight the necessity for studies to adopt AI systems capable of providing personalized feedback to individual learners.

Despite these advancements, current research on LLM-based feedback systems often lacks validation with real-world student data or focuses on proof of concept studies without inviting domain experts as collaborators \citep{nguyen2023evaluating,sessler2023peer,gabbay2024combining}. This gap underscores the need for developing and validating feedback systems that are not only technically sound but also pedagogically effective and aligned with didactic principles.

In science education, experimentation is a fundamental practice essential for fostering analytical thinking and understanding the empirical process of scientific inquiry \citep{NationalResearchCouncil2012}. 
However, students frequently encounter challenges and make mistakes in designing and conducting experiments — such as forming testable hypotheses, controlling variables, and interpreting data \citep{kranz2023learners, baur2018fehler}. As experimentation protocols are the central student outcomes of the experimentation process they are valuable sources for assessment and feedback. Addressing errors in experimentation protocols typically requires personalized feedback from educators, a process that is both time-consuming and resource-intensive.

Responding to these challenges and following the call by \citet{zhai2023ai} for innovation in science education through generative AI, we aim to develop and validate an LLM feedback agent that (1) detects errors in students' experimentation protocols and (2) provides adaptive feedback to learners in real-time. By leveraging LLMs, our system seeks to reduce the workload on educators while enhancing the learning experience for students. Our research addresses a significant gap that still exists between theoretical work on generative AI in science education and practical applications, including a focus on real-world implications.

Building upon our previous work (see: \citet{bewersdorff2023assessing}), where we developed an AI system to automatically detect errors in experimentation protocol, we now extend our approach to deliver actionable feedback that addresses specific student misconceptions and mistakes. We aim to investigate whether LLMs possess the capacity to comprehend context and provide meaningful, adaptive feedback that aligns with pedagogical best practices.

To achieve this, we conducted a study comparing feedback texts on students' experimentation errors written by 11 biology pre- and in-service teachers as well as 5 science education experts  with feedback generated by our LLM agent. Subsequently, 4 raters, professionals in science education and teaching experience, evaluated the feedback texts across six dimensions, including language- and content-related aspects.
By integrating authentic student data and collaborating closely with real-world educators, our research contributes an LLM feedback agent that is validated according to educational needs and didactic principles. 
%
%
In summary, our contributions are:

\begin{enumerate} 
\item The development of an LLM feedback agent to detect logical errors in students' experimentation protocols and provide adaptive feedback.

\item A comprehensive multi-dimensional evaluation analysing the quality of the LLM-generated feedback compared to teacher feedback, validated through content-related and language-related aspects.

\item The application to real-world student experimentation protocols to compare the LLM agent feedback to that of practicing teachers as well as science education experts to demonstrate its practical implementation in educational settings and theoretical soundness.

\end{enumerate}

%% file: 02_theory.tex
\section{Theoretical Background}

\subsection{Characteristics of effective feedback} \label{sec:Characteristics_feedback}

Focusing on the learning process, formative assessment tries to continuously adapt the teaching to the needs of the students \citep{filsecker2012repositioning}. It has been, despite some critics \citep{bennett2010cognitively}, identified as one of the most significant influencing factors for effective learning \citep{hattie2008visible}. 
Feedback is a key part of formative assessment \citep{brookhart2008feedback}, helping to bridge the gap between current understanding and learning goals \citep{sadler1989formative}. \citet{shute2008focus} defined formative feedback as ``information communicated to the learner that is intended to modify his or her thinking or behavior to improve learning''. According to the model of \cite{hattie2007power}, effective feedback answers the following the questions ``Where am I going?'' \textit{(Feed Up)}, ``How am I going?''  \textit{(Feed Back)}, and ``Where to go next?''  \textit{(Feed Forward)}. Each feedback question again can be applied on four levels: The task level, the process level, the self-regulation level and the self level (see \cite{hattie2007power}).

To guide students towards their academic goals, content-related feedback is fundamental. Firstly, \textit{Feed Up} involves clearly summarizing the objectives of the student, establishing a clear goal, and providing a sense of direction \citep{hattie2007power}. 
Secondly, \textit{Feed Back} focuses on the student's current performance, highlighting errors in context. This helps students understand their situation and identify specific areas for improvement \citep{hattie2007power}.
Lastly, \textit{Feed Forward} offers guidance on the next steps. Overall, effective feedback facilitates the learning process, by identifying errors but also recalling the objectives and providing hints for solutions or giving explanations to address identified challenges \citep{hattie2007power, ossenberg2019attributes}.

Next to the content-related factors, the tone, linguistic clarity,  appropriate use of technical terms, and the length of the text play a crucial role in the effectiveness of feedback. 
The tone should be encouraging while avoiding excessive positivity \citep{kluger1996effects}. This approach creates a supportive learning environment and ensures that the feedback remains constructive \citep{brookhart2017give}. A well-balanced tone motivates students while preserving the critical insights essential for their growth.
Scientific language often creates a significant barrier to understanding scientific concepts \citep{gabel1999improving, cassels1983meaning}. Therefore, feedback needs to be specific \citep{moreno2004decreasing}, straightforward, and easily understandable \citep{brookhart2017give, ossenberg2019attributes, Mory2004}. Using simple sentence structures and clear language tailored to students helps keep feedback focused and manageable, avoiding information overload.
Furthermore, using technical terms adequately is essential for effective feedback. While scientific language can enhance learning \citep{fazio2019science}, it can also be challenging for students to understand \citep{vladuvsic2016understanding, hamnell2023scientific}. To balance this, feedback should include relevant terms without overwhelming students, promoting both understanding and educational value.
Overall, regarding the length, feedback should be as simple and focused as possible \citep{kulhavy1985conjoint}. However, \citet{van2015effects} found in their meta-analysis that more elaborate feedback led to higher learning outcomes than (overly) simplified feedback.



In the field of science education, teachers often favor summative assessment for inquiry-based science learning \citep{grob2017formative, zlabkova2024development}. But, in line with findings from other fields research, especially formative feedback can enhance students' understanding in science education \citep{schultz2019effects} and of scientific inquiry, like experimentation \citep{grangeat2024introduction}.

\subsection{Automated computer-based feedback} 

The effectiveness of automated, computer-based feedback is known for decades in educational research, as it shows some benefits to human-generated feedback. One central advantage is the instance generation of feedback. Especially for low-achieving learners this immediate feedback has proven to be effective \citep{mason2001providing}. In their meta-analysis of the effects of feedback in computer-based instruction, \citet{azevedo1995meta} found effects of .80 for immediate feedback and .35 for delayed feedback across all learners. 
Another advantage can be the lower bias that is perceived when using automated feedback \citep{kluger1996effects}. 
However, one-to-one tutoring is generally very effective \citep{bloom19842}, and computer-based feedback can have difficulties mimicking the adaptiveness of human tutoring.
Unlike conventional computer-based feedback systems, LLM-based systems can offer a higher degree of interactivity, thus enabling a more adaptive and engaging learning experience.





\subsection{LLM-based feedback}


LLM-generated feedback has been explored in various educational fields. In mathematics education, \citet{nguyen2023evaluating} evaluated the effectiveness of GPT-3.5 in addressing student errors related to decimal numbers and found it generally effective in providing appropriate feedback. For writing instruction, \citet{sessler2023peer} utilized GPT-3.5 to generate instant feedback, demonstrating its utility in supporting writing development through constructive critiques. A significant focus has been placed on programming education, where several studies have investigated the capabilities of LLMs. \citet{gabbay2024combining} found that while GPT-3.5 and GPT-4 are effective at detecting errors in code assignments within MOOCs, they often fall short in providing accurate or actionable feedback. \citet{estevez2024evaluation} observed limitations in LLMs' ability to evaluate exercises involving concurrency errors, highlighting challenges in understanding complex programming concepts. Additionally, \citet{koutcheme2024open} noted a tendency for GPT-4 to provide overly positive feedback in introductory programming courses, potentially overlooking critical issues that require attention.

While these advancements are promising, research on LLM-based feedback generation in science education is still developing. From a theoretical standpoint, \citet{zhai2023ai} underscore the importance of embracing AI's potential in formative assessment within science education, advocating for constructive dialogue rather than dismissing its impact. In an earlier review, \citet{zhai2020applying} examined machine learning-based science assessments across three axes -technical implementation, validity, and pedagogical features - and found that most studies concentrated on validity, often neglecting technical and pedagogical aspects.

Recent efforts have begun to explore the application of language models in science education assessment. \citet{wu2023matching} applied a pre-trained BERT model with zero-shot prompting to score students' written responses, demonstrating the feasibility of using LLMs for assessment purposes. Similarly, \citet{latif2024fine} compared a fine-tuned BERT model with GPT-3.5 for automatically scoring student answers, finding that GPT-3.5 significantly outperforms BERT in terms of scoring accuracy. However, these studies primarily focus on scoring student responses rather than generating detailed feedback that can guide learning.
\citet{guo2024using} developed a multi-agent system to generate feedback on science education items. However, the study has notable limitations. It does not compare the system’s feedback to that of real-world teachers, leaving its alignment with practical classroom feedback untested. Additionally, the evaluation focuses solely on over-praise and over-inference, neglecting other critical pedagogical aspects such as clarity or specificity. This narrow scope, combined with limited validation for diverse classroom contexts, raises questions about its broader applicability in education.

There remains a significant gap in research regarding the generation of feedback texts for hands-on student experimentation in science education, particularly in benchmarking against real-world teacher feedback and validating critical feedback quality dimensions. Addressing this gap, our study analyzes the performance of LLM-generated feedback on students' experimentation protocols, comparing it to feedback from practicing teachers and science education experts. Our evaluation spans multiple dimensions and incorporates real-world student data.

%% file: 03_methods.tex
\section{Methodology}

In our study, we conduct a multidimensional comparison of feedback generated by our LLM agent for student errors in real-world experimentation protocol to feedback texts given by teachers and experts in science education. In the following, we introduce our used rating scheme, explain the data collection and automated feedback generation and the analyzes we conducted to evaluate the LLM-generated feedback texts.

\subsection{The Research Context: Students' Errors in Experimentation Protocols}

Planning and conducting investigations, along with analyzing and interpreting data, are considered key scientific practices essential for developing scientific literacy \citep{NationalResearchCouncil2012}. These practices are effectively fostered through inquiry-based learning methods, particularly student-centered experimentation. However, while experimentation is central to scientific inquiry, it demands high levels of scientific reasoning and literacy from students.

Students frequently encounter challenges in planning and conducting experiments, which often result in errors. These errors have been well-documented and empirically confirmed in numerous studies \citep{kranz2023learners}. Common errors include formulating hypotheses that focus on expected observations rather than the dependent variable \citep{baur2018fehler} or neglecting to include control trials \citep{dasgupta2014development, germann1996identifying}.

Building on these documented errors, in a previous study, we developed and validated an LLM-based system to identify 16 common student errors related to experimentation \citep{bewersdorff2023assessing}. 
Each protocol included sections for `Hypothesis', `Materials', `Sketch of the Experimental Setup', `Description of the Implementation', `Observation', and `Result'. The protocols were collected from 37 sixth to eighth-grade students attending secondary schools in Germany and are publicly available in the supplementary material of our previous work (see \citet{bewersdorff2023assessing}). 
%
The student protocols were based on two experimental contexts:

\begin{itemize}

\item \textbf{Yeast Experiment} Students were tasked with determining the conditions necessary for yeast to produce carbon dioxide (``Find out what yeast needs to produce carbon dioxide''). 

\item \textbf{Pine Cone Experiment} Students explored the factors that cause pine cone scales to close (``Find out what triggers cone scales to close''). 

\end{itemize}

In total, the 65 experimentation protocols from the two experimental contexts contained 159 errors. We used this dataset as a basis for our study.

\subsection{Feedback collection from human participants}

From the protocol dataset, we selected 40 protocols to create a test dataset containing a total of 109 student errors. The size of the dataset is limited by the inherent challenges of collecting data on student experimentation. Despite its modest size, the dataset was carefully curated to capture a wide range of error types and reflect realistic classroom scenarios \citep{baur2018fehler}. The data was collected from a sample of 37 students in grades six through eight (ages 10–12) across multiple schools. Efforts were made to ensure an approximately equal distribution of genders (boys and girls). Additionally, the sample was designed to ensure diversity regarding performance by including students with high, average, and low performance levels in mathematics, science, and German.

For each error, we collected two human feedback texts to serve as benchmarks for our LLM agent. To achieve this, we engaged 11 science teachers (6 in-service teachers and 5 pre-service teachers in their final semester) and 5 experts in science education who are currently working as university researchers. These two sources provided both real-world feedback from teachers and additional benchmark feedback from domain experts in science education. This selected sample serves as a preliminary foundation for investigating the potential of LLMs in providing feedback on scientific inquiry tasks.

All experts were provided with the student protocol, a list of the identified errors, and general feedback guidelines based on our rating scheme (see Section \ref{sec:rating_scheme}). They also received detailed information about the structure of an experimentation protocol and the tasks assigned to the students. The participation was voluntary, all teachers received a €10 voucher for their support.
%

\begin{figure}[htbp]
    \centering
    \begin{tcolorbox}[colback=gray!5!white, colframe=gray!75!black, width=\textwidth, arc=1mm, auto outer arc, boxrule=0.2mm, title=Prompt for automated feedback generation on experimental protocols]
    \begin{small}
        You are a science teacher. You will give feedback for students in science class that learn about experimentation. \\

        \textless Task\textgreater{}  + \textless Student Protocol\textgreater \\

        \textless Occurred Error\textgreater \\

        Follow the guidelines step by step. \\
        1. Reformulate the problem with your own words specific to the given protocol. Do not just repeat the protocol. Write around 100 words. \\
        2. Take the output of step 1 and answer the question: ``why does this specific protocol contain an error?". Be as concrete as possible. Write a new denser feedback containing all information from 1 and 2. \\
        3. Take the output of step 2, answer the question: ``what could the student do to fix the error?" Be as concrete as possible and stay as close at the original protocol as possible. Write a new denser feedback containing all information from 2 and 3. \\
        4. Reformulate the outcome of step 2 and 3 into a feedback that you as a teacher could give the student. Write here around 200 words. Remember speaking to a student. Be very precise. \\
        5. Reformulate the outcome of step 4 in 100 words. Do not remove any content, just write denser. \\
        
        Return a JSON with keys 1, 2, etc.
    \end{small}
    \end{tcolorbox}
    \caption{Prompt to generate feedback texts on all occurred error in the student experimental protocols.}\label{fig:prompt}
\end{figure}

\subsection{Technical background of the LLM agent}
For automatic feedback generation, we exploited a zero-shot approach, optimizing the prompt for our LLM agent with clear instructions but without requiring prior task-specific examples.
The model was role-prompted to act as a science teacher. For the context, it was then provided with the student's task and the relevant section of the protocol. For example, if the error related to the hypothesis, only the first part of the protocol was included. For errors associated with the result, the entire student-written protocol was provided. The model also received a definition of the specific error identified. Finally, it was instructed to follow a step-by-step approach and return its output in the defined format. The complete prompt is shown in Figure \ref{fig:prompt}.

The prompt was created and refined using 15 protocols selected as train dataset including 50 error of the real-world dataset. The 40 test protocols later used for the comparison with the humans were not considered for the prompt adjustments.
We generated all feedback texts using the OpenAI pipeline, leveraging GPT-3.5 (\texttt{gpt-3.5-turbo-0125}) as baseline LLM. At the time of running the experiments, this version was the most cost-effective option, making it the most practical choice for a classroom setting. 
Notably, since then, the newer GPT-4o Mini model (\texttt{gpt-4o-mini}) has become available, offering even lower costs and improved reasoning capabilities \cite{sessler2024benchmarking}, which could further enhance the feasibility and effectiveness of automated feedback generation in educational contexts.

\subsection{Feedback rating scheme} \label{sec:rating_scheme}

After collecting the feedback tests from humans and our LLM agent, we used a multi-dimensional rating scheme to evaluate them.
Drawing from the theoretical foundations of effective feedback in science education (see \ref{sec:Characteristics_feedback}), we derived a rating scheme based on six dimensions for our analysis. 

According to \citet{hattie2007power}, feedback comprises three core components: \textit{Feed Up}, \textit{Feed Back}, and \textit{Feed Forward}. \textit{Feed Up} clarifies the current goals and learning objectives, ensuring that students understand what is expected of them. \textit{Feed Back} identifies specific errors within the context of the current work, providing students with clear insights into their mistakes. \textit{Feed Forward} suggests possible courses of action to address the identified errors, guiding students on how to improve and proceed with their learning.

In addition to these components, the format of the feedback plays a crucial role in its effectiveness (\ref{sec:Characteristics_feedback}). Therefore, we included three language-related dimensions: \textit{Constructive Tone}, \textit{Linguistic Clarity}, and \textit{Technical Terminology}. A \textit{Constructive Tone} ensures that the feedback is delivered in an encouraging and supportive manner without being overly positive, fostering a positive learning environment. \textit{Linguistic Clarity} involves the use of clear and straightforward language that is appropriate for students in grades 6 to 8, making the feedback easily understandable. Lastly, \textit{Technical Terminology} refers to the appropriate use of subject-specific terms to enhance understanding and ensure that the feedback is both precise and relevant to the scientific context.

Each of these six dimensions was evaluated using a 5-point Likert scale, ranging from \textit{Strongly Disagree} to \textit{Strongly Agree}. This quantitative approach enables us to systematically rate and compare the quality of feedback provided by both human experts in the field of science education and the LLM agent. Table \ref{tab:criteria} provides an overview of the dimensions and their corresponding descriptions.

\begin{table}[htbp]
    \centering
    \caption{Overview of the multidimensional criteria applied to assess the feedback, categorized into content-related (C) and language-related (L) aspects.}
    \label{tab:criteria}
    \begin{tabular}{lcl}
    \toprule
    Criteria & Type & Description  \\
    \midrule
         Feed Up & C & Summarizes the goal to be achieved  \\
         Feed Back & C & Describes the error in the context of the current situation  \\
         Feed Forward & C & Indicates solutions or explains methods to address errors   \\
         Constructive Tone & L & Encouraging without being overly positive  \\
         Linguistic Clarity & L & Uses clear sentence structures suitable for grades 6–8  \\
         Technical Terminology & L & Appropriately employs technical terms for grades 6–8  \\
    \bottomrule
    \end{tabular}
    
\end{table}

In addition to the rating scheme, we also measured the length of the feedback text (i.e. the number of words) to analyze its comprehensiveness and conciseness, since it is important to not overwhelm the students by too verbose feedback \citep{kulhavy1985conjoint}.

\subsection{Expert feedback ratings}

After the data collecting phase, we presented the three feedback texts (generated by teacher, expert and LLM agent) blinded and shuffled to four human raters, that are professionals in science inquiry as well as in teaching and learning. The criteria for selection as an rater included: 1) a PhD in a natural science or science education, 2) several years of teaching experience with pre-service science teachers, and 3) publications on experimentation as a method of scientific inquiry. 
The raters were provided with a batch of the original protocols and the task to the assess the texts based on the six aspects in Table \ref{tab:criteria}. 


\subsection{Analysis}

To thoroughly compare the feedback texts written by teachers, experts, and the LLM agent, we conducted several analyses, using human-generated feedback as the benchmark for evaluating the agent's performance. First, we assessed the overall performance by averaging all six categories of the rating scheme into a single overall score and analyzing the distribution of these scores across the three groups, providing a general impression of performance (Section \ref{sec:overall_scoring}).

Next, we conducted a detailed aspect-level analysis to examine each dimension of the rating scheme individually. By comparing the mean and variance of scores for each category, we identified significant differences among the feedback sources—teachers, experts, and agent—using independent t-tests (Section \ref{sec:multidim}).
Since feedback length is also a critical factor for adequacy, particularly because LLMs tend to produce verbose responses, we analyzed the number of words in each feedback text to investigate this possible challenge (Section \ref{sec:length}).

To explore the alignment between feedback from the LLM agent, teachers, and experts, we performed correlation analyzes using Spearmans correlation coefficient $\rho$. This helped us determine whether they excelled or struggled with the same texts or showed strengths on distinct examples (Section \ref{sec:corr_analysis}).
Finally, to gain deeper insight into the challenges faced by the AI system, we closely examined an illustrative example where the LLM agent failed to provide valuable feedback while humans succeeded. This qualitative analysis highlights specific areas where the AI system still lags behind, providing a clearer understanding of the remaining issues (Section \ref{sec:qualitative_example}).

%% file: 04_results.tex
\section{Results}

In following we report the results of the ratings in the multi-dimensional criteria to analyze the feedback quality of our LLM agent compared to real-world teachers and experts. We analyze how well LLMs perform in terms of content and language aspects in providing feedback on experimental protocols.

\begin{figure}[htbp]
    \centering
    \includegraphics[width=0.5\textwidth]{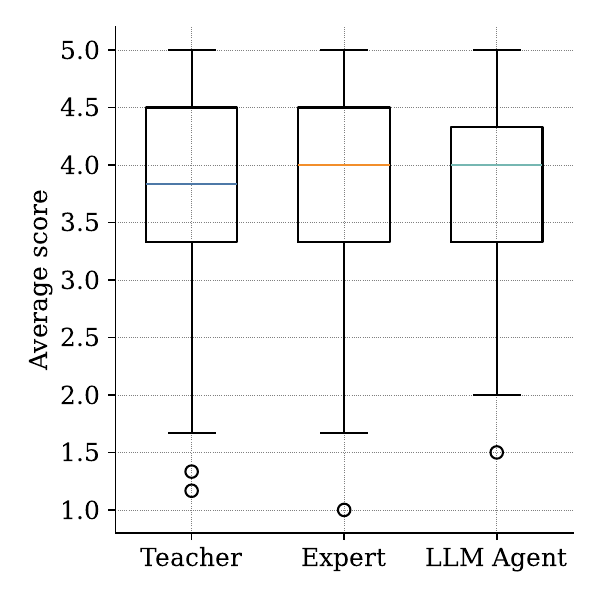}
    \caption{Distribution of the rating scores of teachers, experts and LLM agent averaged across all six dimensions on feedback quality.}
    \label{fig:average_scoring}
\end{figure}

\subsection{Overall scoring}\label{sec:overall_scoring}

As a first step, we compare the mean score, averaging all six multi-dimensional aspects for each feedback text. Figure \ref{fig:average_scoring} picture the distribution for each group. On average, the LLM agent reaches a mean value of 3.784 ($SD=1.238$), the teacher feedback texts 3.805 ($SD=1.266$) and the experts 3.831 ($SD=1.287$), with no significant difference between the three groups. Therefore, from an overall perspective, the feedback generated by an LLM agent can be considered similar quality to the one written by a human teacher or expert.

\subsection{Multidimensional scoring} \label{sec:multidim}

\begin{figure}[htbp]
    \centering
    \includegraphics[width=0.9\textwidth]{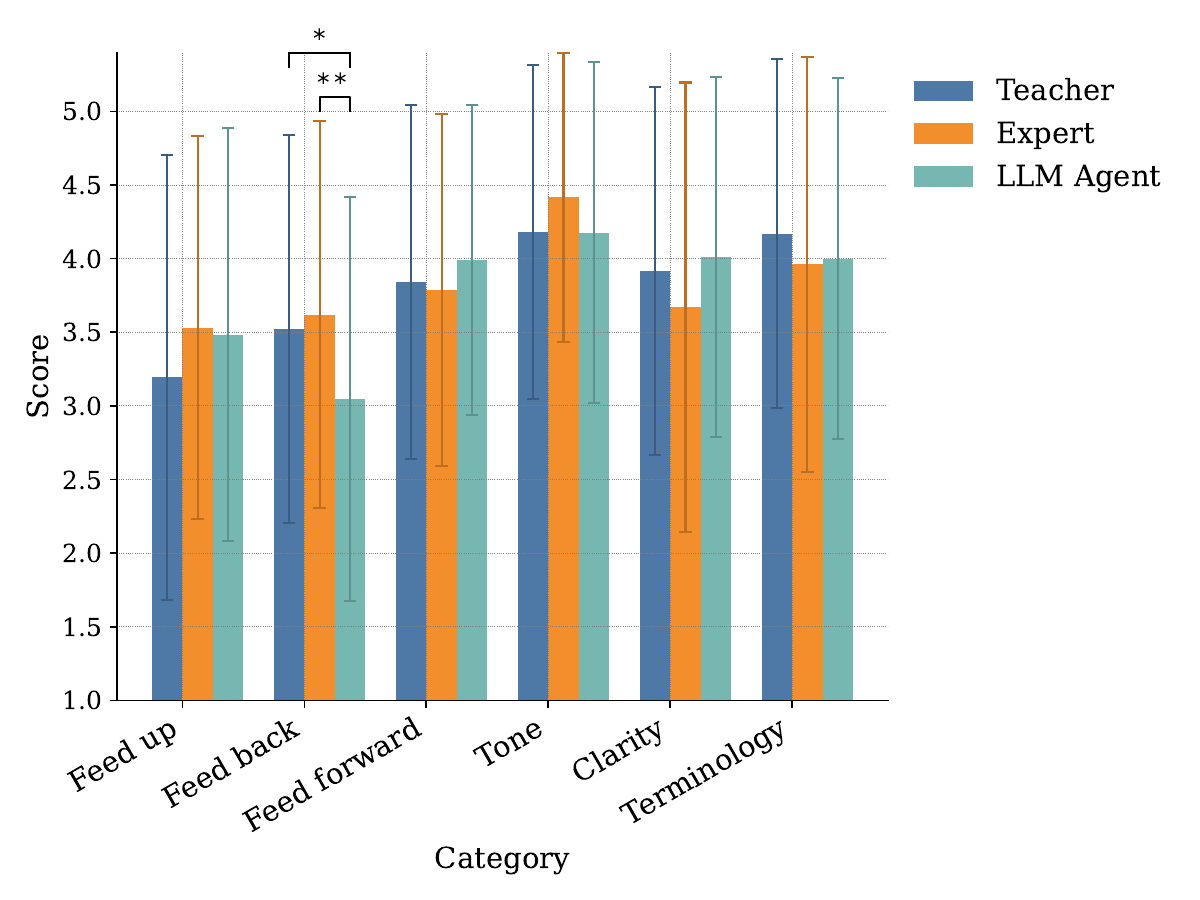}
    \caption{Average score and standard deviation of the scoring of the feedback texts generated by teachers, experts and LLM agent in each rating category. Significant differences are marked by $*$ ($p<0.05$) and $**$ ($p<0.01$).}
    \label{fig:scores_categories}
\end{figure}

\begin{table}[htbp]
    \centering
    \begin{tabular}{lcccccc}
    \toprule
    & Feed Up & Feed Back & Feed Forward & Tone & Clarity & Terminology \\
    \midrule
Teacher & 3.20 $\pm$ 1.51  & 3.52 $\pm$ 1.32  & 3.84 $\pm$ 1.20  & 4.18 $\pm$ 1.13  & 3.91 $\pm$ 1.25  & 4.17 $\pm$ 1.18  \\ 
Expert & 3.53 $\pm$ 1.30  & 3.62 $\pm$ 1.31  & 3.78 $\pm$ 1.19  & 4.42 $\pm$ 0.98  & 3.67 $\pm$ 1.53  & 3.96 $\pm$ 1.41  \\ 
LLM & 3.48 $\pm$ 1.40  & 3.05 $\pm$ 1.37  & 3.99 $\pm$ 1.05  & 4.18 $\pm$ 1.16  & 4.01 $\pm$ 1.22  & 4.00 $\pm$ 1.22  \\ 
\bottomrule
    \end{tabular}
    \caption{Average score and standard deviation of the ratings of the feedback texts for all three groups.}
    \label{tab:scores_categories}
\end{table}

To get more detailed insights, we analyze the ratings of the feedback texts in the multi-dimensional aspects. Here, Figure \ref{fig:scores_categories} depicts the mean and standard deviation of all the categories. In general, all three groups achieved a higher rating in the three language related aspects and slightly worse results on the content-related ones. The details can be found in Table \ref{tab:scores_categories}.

Investigating the language categories, we see that while experts seem to have a more constructive tone, they suffer more in linguistic clarity. Teachers seem to have a small advantage using technical terminology for the young target group. But, overall the variations are minor and there are no significant differences between the three groups. Therefore, from a linguistic point of view, an LLM agent is able to provide adequate feedback for students from grade 6 to 8.

Considering the content-related aspects, we see more variations between the raters. Teachers tend to have more struggle clarifying the current goal of the experiment, and the LLM agent has a small advancement suggesting possible future actions. But, the only significant difference ($p<0.05$) is found for \textit{feed back}, where teachers and experts achieve a better performance then the LLM. This indicates that the LLM agent has still difficulties identifying the error within its context and providing insights into the mistake. For writing the \textit{Feed back} part, the model needs to completely understand the step taken by the student and interpret them correctly. This is particular tricky due to the incomplete and unstructured written protocols written by the students. Here, teachers and expert may have more experience in understanding the context and therefore providing better tailored feedback. 

\subsection{Length analysis}\label{sec:length}

\begin{figure}[htbp]
    \centering
    \includegraphics[width=0.8\textwidth]{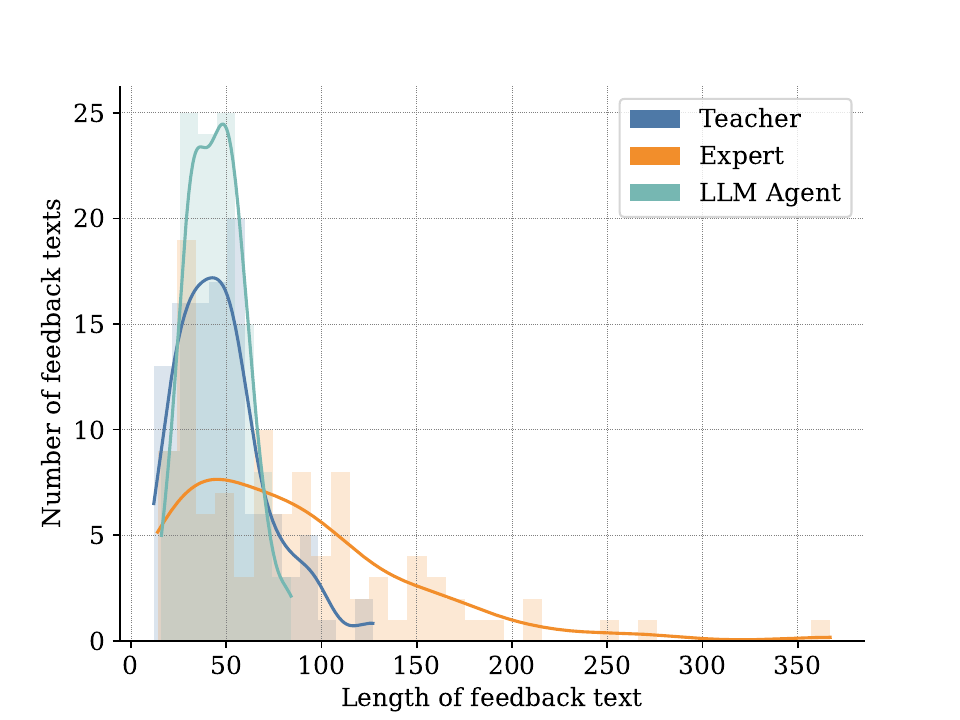}
    \caption{Distribution of the number of words in each feedback text written by the teachers, experts and the LLM agent.}
    \label{fig:length_distribution}
\end{figure}

A good feedback need to be comprehensive as well as precise. Therefore, we take a look into the length of the provided feedbacks of the three groups in Figure \ref{fig:length_distribution}. In the prompt for the LLM agent we asked to write 100 words, interestingly most generated feedbacks contain 50 words. This aligns with the length of the formative feedback provided by the teachers and is therefore probably a realistic length for a real-world classroom. On the other hand, the experts provided clearly more lengthy feedback texts, which is not suitable for students of grade 6 to 8 anymore. Even though their text are superior in form of tone and feed back content, the real-world application is questionable. Here, the LLM agent can be a consistent source of feedback of suitable length.

\subsection{Correlation analysis}\label{sec:corr_analysis}

To assess whether humans and LLMs excel or fall short for similar feedback texts or struggle with different kinds of errors, we analyzed the correlations of the expert ratings.

Correlations for language-related aspects were generally strong. \textit{Constructive Tone} showed moderate to strong correlations across all constellations ($\rho \geq .48$), reflecting meaningful alignment. \textit{Linguistic Clarity} demonstrated robust agreement across all sources ($\rho \geq .60$), while \textit{Technical Terminology} exhibited moderate positive correlations ($\rho \geq .41$). Detailed correlation values are provided in Table \ref{tab:correlations}.

In contrast, content-related aspects showed consistently weaker correlations. Between experts and teachers, correlations ranged from weak to moderate ($\rho = .26$ for \textit{Feed Forward} to $\rho = .41$ for \textit{Feed Back}), suggesting some overlap in performance on distinct feedback texts. However, correlations between human-written feedback and LLM-generated feedback were low to non-existent. For example, correlations between teachers and the LLM ranged from $\rho = .06$ for \textit{Feed Back} to $\rho = .21$ for \textit{Feed Up}, while correlations between expert feedback ratings and LLM feedback ratings were negligible ($|\rho| < .1$ for all aspects). Table \ref{tab:correlations} provides an overview over all correlation values. These results suggest that humans and LLMs struggled with different types of feedback texts, with limited alignment in their strengths.

\begin{table}[ht]
    \centering
    \begin{tabular}{lccc}
        \toprule
        & Teacher-Expert & Teacher-LLM & Expert-LLM \\
        \midrule
        Feed Up           &      .36      &    .21       &     -.05      \\
        Feed Back         &     .41       &    .06       &    .07       \\
        Feed Forward      &      .26      &     .18      &    .01       \\
        Constructive Tone &       .59     &    .54       &    .48       \\
        Linguistic Clarity&     .60       &     .61      &    .61       \\
        Technical Terminology &   .45    &     .48      &      .41     \\
        \bottomrule
    \end{tabular}
    \caption{Comparison of the Spearman correlations coefficients $\rho$ across all six rating dimensions}
    \label{tab:correlations}
\end{table}

\subsection{Qualitative analysis}\label{sec:qualitative_example}

To gain deeper insight into why our LLM agent performs sometimes worse than the humans, we qualitatively compared the feedback provided by these three entities regarding a specific student error. The error analyzed in Table \ref{tab:qualitative_example} was selected based on two criteria: (1) poor performance by the LLM agent on content-related aspects, and (2) the largest overall gap between human and LLM feedback across all dimensions. The second criterion highlights cases where humans excel at providing effective feedback while the LLM agent falls short.

\newcommand{\feedup}[1]{\textcolor{orange}{#1}}
\newcommand{\feedback}[1]{\textcolor{purple}{#1}}
\newcommand{\feedforward}[1]{\textcolor{blue}{#1}}

\begin{table}[htbp]
    \centering
    
\begin{tabular}{p{1cm}p{8.3cm}P{1.2cm}} 
\toprule
\textbf{} & \textbf{Feedback Text} & \textbf{Ratings} \\ \midrule
Expert         & \feedup{With a hypothesis, you should set a goal for your research project that you can confirm or refute with your experiment/trial.} Pay attention to the points for formulating a hypothesis that we discussed in class. Keep in mind that the hypothesis should be confirmable or refutable. \feedback{In your example, you have mixed various variables so that one cannot make distinctions and cannot confirm or refute it in that form}. Ensure that your hypothesis is clearly defined and all parts of your hypothesis are verifiable. For example: is it really temperature AND without light, or is it and/or? \feedforward{Try again to separate your variables and structure one or possibly two hypotheses.}   & FU: 5.0 FB: 5.0 FF: 5.0     \\  
Teacher      & \feedback{Your hypothesis should not contain all your assumptions.} Initially, \feedforward{limit yourself to one, for example, the influence of temperature}. Based on these results, you can formulate a new hypothesis. Your hypothesis contains the dependent variable, but try to formulate a complete German sentence.      & FU: 1.0 FB: 5.0 FF: 4.0    \\ 
LLM     & \feedforward{Concentrate in your assumption on one variable to identify the trigger for the closure of the cone scales.} Conduct separate experiments for cold temperatures and darkness to determine the specific factor. Keep the variables controlled and constant to achieve accurate results.     & FU: 1.0 FB: 1.0 FF: 5.0         \\ \bottomrule 
\end{tabular}

    \caption{Illustrative example for the different feedback texts written by Science Education Experts, Teachers and LLM agent. The original hypothesis by the student was: \textit{That at low temperatures and without light shining in, the pine cone closes}. Here, the identified error is: \textit{Hypothesis consists of a combination of variables}. The Feedback texts are rated in the six categories, we colored the relevant parts of the texts including \feedup{Feed Up (FU)}, \feedback{Feed Back (FB)} and \feedforward{Feed Forward (FF)}. For the publication of this article the texts were translated from German into English.} 
    \label{tab:qualitative_example}
\end{table}

The expert provided comprehensive feedback with high scores across all dimensions. It repeats the objective of formulating a  hypothesis (``set a goal [...] that you can confirm or refute''), identifies the status of the student's hypothesis and its problem (``mixed various variables'', ``cannot confirm or refute in that form'') and offers a clear suggestion on how to proceed (``separate your variables''). While the text includes all relevant aspects of good feedback, it is notably lengthy, consisting of 112 words - approximately twice the ideal length for students of this age group.
The teacher's feedback is generally strong but lacks initial context, jumping directly to the problem with the hypothesis (``should not contain all your assumptions'') before providing guidance on how to improve it (``limit yourself to one''). At 43 words, the feedback is concise and appropriately tailored for the target group.
The LLM agent's feedback, consisting of 41 words, focuses on explaining how to improve the experiment and what steps to take next (``concentrate your assumption on one variable''). However, it fails to establish a shared understanding of the goal of the hypothesis and does not explain the resulting problem with the student's hypothesis(e.g., why the current approach is difficult to test).

All three texts scored 5 points for appropriate use of technical terminology and 4 points for constructive tone. However, linguistic clarity varied: the expert received 3 points, the teacher 2, and the LLM agent only 1. The LLM agent's feedback remained high-level, using vague expressions like "specific factor" and failing to clarify how to keep variables "controlled and constant," which limited its usability for students.


This example highlights that while the LLM agent effectively communicates with clarity comparable to human feedback and provides actionable guidance on next steps (e.g., focusing on one variable), it struggles to identify and articulate the specific problem in the student’s experimentation protocol and contextualize it effectively. 
This shortage may stem from several factors. Generally, student data tends to be very noisy, and in this case, the phrasing of the hypothesis is highly unusual, which can make it challenging for the model to fully comprehend the underlying idea. Furthermore, this specific type of student data is quite rare and likely not part of the model's training data. As a result, the model may lack experience in handling and generating this kind of data. Additionally, perfect feedback—encompassing feed up, feed back, and feed forward—might also be underrepresented in the training data. It is possible that many ``feedback'' texts focus primarily on next steps while giving less attention to feed up and feed back. Referring back to Section \ref{sec:multidim}, the LLM agent performs significantly better in addressing the feed forward component compared to the other two parts.
%

These qualitative findings highlight that while LLM agent can provide valuable guidance on next steps and use appropriate technical language, there are still negative examples where they currently lag behind teachers and human experts in delivering contextual understanding and clear communication.


%% file: 05_discussion.tex
\section{Discussion}
The present study evaluated the efficacy of an LLM feedback agent on experimentation protocols compared to human experts and teachers across multiple dimensions of feedback quality: \textit{Feed Up}, \textit{Feed Back}, \textit{Feed Forward}, \textit{Constructive Tone}, \textit{Linguistic Clarity}, and \textit{Technical Terminology}. Our findings indicate that the LLM agent performed on par with human experts and teachers in most dimensions, showing no significant differences. However, it was significantly less effective in the \textit{Feed Back} dimension, which involves describing the error in the context of the current situation.

\paragraph{LLMs can provide actionable and linguistically appropriate feedback.} The comparable performance of the LLM agent in the content-related dimensions such as \textit{Feed Up}, \textit{Feed Forward} as well as the language-related dimensions shows that an LLM can effectively mimic feedback from educational experts in summarizing goals, providing forward-looking guidance, maintaining an encouraging tone, and using appropriate language and terminology for middle school students. This aligns with previous research indicating that AI systems can deliver structured and pedagogically sound feedback \citep{guo2024using, sessler2023peer,fung2024automatic,cohn2023towards}.

\paragraph{LLMs fall short in nuanced context understanding.}
The small but significant shortfall in the \textit{Feed Back} dimension highlights a critical area where LLM agents need improvement. Describing errors within the specific context of a student's work requires a nuanced understanding of the learning material and the student's thought process. Human teachers excel in this area due to their ability to interpret subtleties in student responses and provide feedback that addresses individual misconceptions \citep{shute2008focus}. The LLM agent's deficiency here may stem from limitations in its ability to comprehend context or infer underlying misconceptions from student errors.
However, this performance appears to be typical of current LLM agents, as other studies report that LLM-based systems tend to retain naive or limited pedagogical content knowledge \cite{chapagain2024explanations, tseng2024effectiveness}.


\paragraph{Implementation of AI in the classroom}

The analysis of content-related aspects reveals notable differences between human and LLM-generated feedback, indicating that humans and LLMs excel in distinct error scenarios. While their overall performance is comparable (see Section \ref{sec:multidim}), this suggests that integrating LLMs together with human assessments could enhance feedback quality by combining the strengths of teachers and the LLM agent.

The design of the LLM agent as a tool for hybrid intelligence aligns with the principle of augmenting human capabilities in the classroom rather than replacing them \citep{molenaar2022towards}. Teachers can integrate the LLM agent's strengths into their practice without making big changes to their teaching methods. By supporting adaptive and instant assessment and feedback for all students, the AI system enables teachers to focus on personalized instruction while retaining control over the learning process.

Despite the promising results regarding LLM-generated feedback quality, \citet{nazaretsky2024ai} report that students tend to prefer human feedback due to perceived deficiencies in the genuineness, usefulness, and objectivity of AI feedback. A \textbf{teacher-in-the-loop} approach could address these issues by ensuring quality control and enhancing trust. There is a spectrum ranging from student-centered use of AI, through teacher-only use, to the complete absence of AI in the classroom \citep{molenaar2022towards}. Finding the right level of AI integration into classroom settings remains challenging and is subject to future research. Ideally, these LLM agents are designed to provide different levels of AI integration, providing teachers with options to choose the level that best aligns with their pedagogical goals. For effective AI integration there is a need to build AI literacy among STEM teachers, enabling them to make informed pedagogical decisions based on an understanding of the strengths and limitations of current LLM agents \citep{hornberger2023university}.

\subsection{Limitations}

One major limitation of this study is the small sample size for many of the identified errors, with some errors represented by only a single instance. This restricts the generalizability of our findings and may not capture the full range of possible feedback scenarios. 
Additionally, the LLM capabilities were constrained by financial considerations essential for creating a real-world, school-usable system. As a result, we utilized the cost-efficient version of GPT at the time of conducting the user study, GPT-3.5, which possess significantly lower reasoning abilities compared to more advanced, but expensive, models like GPT-4 or o1. Also, the GPT-4o Mini model (published after running our experiments) is more cost-efficient and shows improved capabilities \citep{sessler2024benchmarking}. 
Choosing this version likely impacted the agent's performance, particularly in areas requiring deeper analytical skills. It is to be expected that newer models are more powerful in understanding context and providing adapted feedback. 
Furthermore, the feedback generated was exclusively based on the students' protocols, which means that both LLM and human feedback might have been limited by the absence of further contextual information. This lack of context could have hindered the ability of human reviewers to provide an even more informed and nuanced feedback.

\subsection{Future work}

Future work should involve conducting a real-world study where the LLM agent is implemented in classroom settings to evaluate its effectiveness and practicality in everyday educational environments and assess the effectiveness beyond the proof-of-concept given in this study. Additionally, upgrading the agent to more advanced models, such as GPT-4o or o1, would likely enhance the agent’s reasoning capabilities as shown by the findings of current research \citep{sessler2024can, latif2024systematic}, allowing for more sophisticated and accurate feedback. 

To address limitations related to sample size, strategies such as data augmentation using generative AI \citep{kieser2023educational} could be employed to increase the number of samples. These methods would allow for more robust findings without the challenging task of collecting large numbers of student protocols in real-world classroom settings.
Importantly, synthetic data generation can be designed to be privacy-preserving by retaining the statistical properties of the original dataset while ensuring that no individual student’s data is directly replicated \citep{vie2022privacy}. Such approaches not only expand the dataset and increase the reliability but also align with ethical considerations and data privacy regulations, which are critical in educational research.

Currently, our LLM agent is limited to text-based interactions. At its core, science learning, and experimentation protocols, like science itself are inherently multimodal \citep{lemke_literacies_2004, kress_vanleeuwen_1990}. Regarding experimentation protocols, these include reading and writing scientific arguments and explanations, drawing and interpreting diagrams, analyzing and visualizing data, and creating flowcharts and diagrams.  
Future systems could adopt a multi-modal approach, such as providing feedback on experimental drawings or generating sketches of experimental designs, thereby addressing the diverse modalities inherent in scientific inquiry \citep{lee2023multimodality, bewersdorff2024taking}.

%% file: 06_conclusion.tex
\section{Conclusion}

We conducted a study to compare feedback on experimentation protocols written by teachers, science education experts, and an LLM agent. Human raters assessed the feedback across six dimensions to evaluate its quality in both content and language-related aspects. Our findings show that, on average, there is no significant difference in overall quality between the three groups, indicating that LLMs are capable of producing valuable feedback for students.
However, the LLM agent tends to focus on the \textit{feed forward} aspect of feedback, partly neglecting the equally important \textit{feed back} component. This highlights the need for further refinement of LLMs to ensure they provide balanced and comprehensive feedback for students. Additionally, since teachers and LLMs appear to encounter challenges with different types of student errors, integrating feedback from both could offer a promising approach for enhancing educational practices in the future.

%% file: main.bbl
\begin{thebibliography}{64}
\providecommand{\natexlab}[1]{#1}
\providecommand{\url}[1]{{#1}}
\providecommand{\urlprefix}{URL }
\providecommand{\doi}[1]{\url{https://doi.org/#1}}
\providecommand{\eprint}[2][]{\url{#2}}
 \bibcommenthead

\bibitem[{Azevedo and Bernard(1995)}]{azevedo1995meta}
Azevedo R, Bernard RM (1995) A meta-analysis of the effects of feedback in
  computer-based instruction. Journal of Educational Computing Research
  13(2):111--127

\bibitem[{Baur(2018)}]{baur2018fehler}
Baur A (2018) Fehler, fehlkonzepte und spezifische vorgehensweisen von
  sch{\"u}lerinnen und sch{\"u}lern beim experimentieren. Zeitschrift f{\"u}r
  Didaktik der Naturwissenschaften 24(1):115--129

\bibitem[{Bennett(2010)}]{bennett2010cognitively}
Bennett RE (2010) Cognitively based assessment of, for, and as learning (cbal):
  A preliminary theory of action for summative and formative assessment.
  Measurement 8(2-3):70--91

\bibitem[{Bewersdorff et~al(2023)Bewersdorff, Se{\ss}ler, Baur, Kasneci, and
  Nerdel}]{bewersdorff2023assessing}
Bewersdorff A, Se{\ss}ler K, Baur A, et~al (2023) Assessing student errors in
  experimentation using artificial intelligence and large language models: A
  comparative study with human raters. Computers and Education: Artificial
  Intelligence 5:100177

\bibitem[{Bewersdorff et~al(2024)Bewersdorff, Hartmann, Hornberger, Se{\ss}ler,
  Bannert, Kasneci, Kasneci, Zhai, and Nerdel}]{bewersdorff2024taking}
Bewersdorff A, Hartmann C, Hornberger M, et~al (2024) Taking the next step with
  generative artificial intelligence: The transformative role of multimodal
  large language models in science education. arXiv preprint arXiv:240100832

\bibitem[{Bloom(1984)}]{bloom19842}
Bloom BS (1984) The 2 sigma problem: The search for methods of group
  instruction as effective as one-to-one tutoring. Educational researcher
  13(6):4--16

\bibitem[{Brookhart(2017)}]{brookhart2017give}
Brookhart S (2017) How to give effective feedback to your students. ASCD

\bibitem[{Brookhart(2007)}]{brookhart2008feedback}
Brookhart SM (2007) Feedback that fits. Educational Leadership 65(4):54--59

\bibitem[{Cassels and Johnstone(1983)}]{cassels1983meaning}
Cassels J, Johnstone A (1983) The meaning of words and the teaching of
  chemistry. Education in chemistry 20(1):10--11

\bibitem[{Chapagain et~al(2024)Chapagain, Sajib, Prodan, and
  Rus}]{chapagain2024explanations}
Chapagain J, Sajib MI, Prodan R, et~al (2024) A study of llm generated
  line-by-line explanations in the context of conversational program
  comprehension tutoring systems. In: Ferreira~Mello R, Rummel N, Jivet I,
  et~al (eds) Technology Enhanced Learning for Inclusive and Equitable Quality
  Education. Springer Nature Switzerland, Cham, pp 64--74

\bibitem[{Cohn et~al(2023)Cohn, Hutchins, and Biswas}]{cohn2023towards}
Cohn C, Hutchins N, Biswas G (2023) Towards a formative feedback generation
  agent: Leveraging a human-in-the-loop, chain-of-thought prompting approach
  with llms to evaluate formative assessment responses in k-12 science.

\bibitem[{Cordova and Lepper(1996)}]{cordova1996intrinsic}
Cordova DI, Lepper MR (1996) Intrinsic motivation and the process of learning:
  Beneficial effects of contextualization, personalization, and choice. Journal
  of educational psychology 88(4):715

\bibitem[{Council(2012)}]{NationalResearchCouncil2012}
Council NR (2012) A Framework for K-12 Science Education: Practices,
  Crosscutting Concepts, and Core Ideas. The National Academies Press,
  Washington, DC

\bibitem[{Dasgupta et~al(2014)Dasgupta, Anderson, and
  Pelaez}]{dasgupta2014development}
Dasgupta AP, Anderson TR, Pelaez N (2014) Development and validation of a
  rubric for diagnosing students’ experimental design knowledge and
  difficulties. CBE—Life Sciences Education 13(2):265--284

\bibitem[{Est{\'e}vez-Ayres et~al(2024)Est{\'e}vez-Ayres, Callejo,
  Hombrados-Herrera, Alario-Hoyos, and Delgado~Kloos}]{estevez2024evaluation}
Est{\'e}vez-Ayres I, Callejo P, Hombrados-Herrera M{\'A}, et~al (2024)
  Evaluation of llm tools for feedback generation in a course on concurrent
  programming. International Journal of Artificial Intelligence in Education pp
  1--17

\bibitem[{Fazio and Gallagher(2019)}]{fazio2019science}
Fazio X, Gallagher TL (2019) Science and language integration in elementary
  classrooms: Instructional enactments and student learning outcomes. Research
  in Science Education 49:959--976

\bibitem[{Filsecker and Kerres(2012)}]{filsecker2012repositioning}
Filsecker M, Kerres M (2012) Repositioning formative assessment from an
  educational assessment perspective: A response to dunn \& mulvenon (2009).
  Practical Assessment, Research \& Evaluation 17(16):n16

\bibitem[{Fung et~al(2024)Fung, Wong, and Tan}]{fung2024automatic}
Fung SCE, Wong MF, Tan CW (2024) Automatic feedback generation on k-12
  students' data science education by prompting cloud-based large language
  models. In: Proceedings of the Eleventh ACM Conference on Learning@ Scale, pp
  255--258

\bibitem[{Gabbay and Cohen(2024)}]{gabbay2024combining}
Gabbay H, Cohen A (2024) Combining llm-generated and test-based feedback in a
  mooc for programming. In: Proceedings of the Eleventh ACM Conference on
  Learning@ Scale, pp 177--187

\bibitem[{Gabel(1999)}]{gabel1999improving}
Gabel D (1999) Improving teaching and learning through chemistry education
  research: A look to the future. Journal of Chemical education 76(4):548

\bibitem[{Gan et~al(2021)Gan, An, and Liu}]{gan2021teacher}
Gan Z, An Z, Liu F (2021) Teacher feedback practices, student feedback
  motivation, and feedback behavior: how are they associated with learning
  outcomes? Frontiers in psychology 12:697045

\bibitem[{Germann et~al(1996)Germann, Aram, and Burke}]{germann1996identifying}
Germann PJ, Aram R, Burke G (1996) Identifying patterns and relationships among
  the responses of seventh-grade students to the science process skill of
  designing experiments. Journal of Research in Science Teaching: The Official
  Journal of the National Association for Research in Science Teaching
  33(1):79--99

\bibitem[{Grangeat et~al(2024)Grangeat, Harrison, and
  Dolin}]{grangeat2024introduction}
Grangeat M, Harrison C, Dolin J (2024) Introduction: Exploring assessment in
  stem inquiry learning classrooms. In: Developing Formative Assessment in STEM
  Classrooms. Routledge, p 1--17

\bibitem[{Grob et~al(2017)Grob, Holmeier, and Labudde}]{grob2017formative}
Grob R, Holmeier M, Labudde P (2017) Formative assessment to support
  students’ competences in inquiry-based science education. Interdisciplinary
  Journal of Problem-Based Learning 11(2):6

\bibitem[{Guo et~al(2024)Guo, Latif, Zhou, Huang, and Zhai}]{guo2024using}
Guo S, Latif E, Zhou Y, et~al (2024) Using generative ai and multi-agents to
  provide automatic feedback. arXiv e-prints pp arXiv--2411

\bibitem[{Hamnell-Pamment(2023)}]{hamnell2023scientific}
Hamnell-Pamment Y (2023) Scientific language use and sensemaking in concept
  maps: Interaction between concept systems, scientific concepts and everyday
  concepts. Knowledge Management \& E-Learning 15(3):448--467

\bibitem[{Hattie(2008)}]{hattie2008visible}
Hattie J (2008) Visible learning: A synthesis of over 800 meta-analyses
  relating to achievement. routledge

\bibitem[{Hattie and Timperley(2007)}]{hattie2007power}
Hattie J, Timperley H (2007) The power of feedback. Review of educational
  research 77(1):81--112

\bibitem[{Hornberger et~al(2023)Hornberger, Bewersdorff, and
  Nerdel}]{hornberger2023university}
Hornberger M, Bewersdorff A, Nerdel C (2023) What do university students know
  about artificial intelligence? development and validation of an ai literacy
  test. Computers and Education: Artificial Intelligence 5:100165

\bibitem[{Kasneci et~al(2023)Kasneci, Se{\ss}ler, K{\"u}chemann, Bannert,
  Dementieva, Fischer, Gasser, Groh, G{\"u}nnemann, H{\"u}llermeier
  et~al}]{kasneci2023chatgpt}
Kasneci E, Se{\ss}ler K, K{\"u}chemann S, et~al (2023) Chatgpt for good? on
  opportunities and challenges of large language models for education. Learning
  and individual differences 103:102274

\bibitem[{Kieser et~al(2023)Kieser, Wulff, Kuhn, and
  K{\"u}chemann}]{kieser2023educational}
Kieser F, Wulff P, Kuhn J, et~al (2023) Educational data augmentation in
  physics education research using chatgpt. Physical Review Physics Education
  Research 19(2):020150

\bibitem[{Van~der Kleij et~al(2015)Van~der Kleij, Feskens, and
  Eggen}]{van2015effects}
Van~der Kleij FM, Feskens RC, Eggen TJ (2015) Effects of feedback in a
  computer-based learning environment on students’ learning outcomes: A
  meta-analysis. Review of educational research 85(4):475--511

\bibitem[{Kluger and DeNisi(1996)}]{kluger1996effects}
Kluger AN, DeNisi A (1996) The effects of feedback interventions on
  performance: a historical review, a meta-analysis, and a preliminary feedback
  intervention theory. Psychological bulletin 119(2):254

\bibitem[{Koutcheme et~al(2024)Koutcheme, Dainese, Sarsa, Hellas, Leinonen, and
  Denny}]{koutcheme2024open}
Koutcheme C, Dainese N, Sarsa S, et~al (2024) Open source language models can
  provide feedback: Evaluating llms' ability to help students using
  gpt-4-as-a-judge. In: Proceedings of the 2024 on Innovation and Technology in
  Computer Science Education V. 1. Association for Computing Machinery, New
  York, NY, USA, p 52–58

\bibitem[{Kranz et~al(2023)Kranz, Baur, and M{\"o}ller}]{kranz2023learners}
Kranz J, Baur A, M{\"o}ller A (2023) Learners’ challenges in understanding
  and performing experiments: a systematic review of the literature. Studies in
  Science Education 59(2):321--367

\bibitem[{Kress and van Leeuwen(1990)}]{kress_vanleeuwen_1990}
Kress G, van Leeuwen T (1990) Reading Images. Deakin University Press, Deakin

\bibitem[{Kulhavy et~al(1985)Kulhavy, Lee, and Caterino}]{kulhavy1985conjoint}
Kulhavy RW, Lee JB, Caterino LC (1985) Conjoint retention of maps and related
  discourse. Contemporary educational psychology 10(1):28--37

\bibitem[{Latif and Zhai(2024)}]{latif2024fine}
Latif E, Zhai X (2024) Fine-tuning chatgpt for automatic scoring. Computers and
  Education: Artificial Intelligence 6:100210

\bibitem[{Latif et~al(2024)Latif, Zhou, Guo, Gao, Shi, Nayaaba, Lee, Zhang,
  Bewersdorff, Fang et~al}]{latif2024systematic}
Latif E, Zhou Y, Guo S, et~al (2024) A systematic assessment of openai
  o1-preview for higher order thinking in education. arXiv preprint
  arXiv:241021287

\bibitem[{Lee et~al(2023)Lee, Shi, Latif, Gao, Bewersdorff, Nyaaba, Guo, Wu,
  Liu, Wang et~al}]{lee2023multimodality}
Lee GG, Shi L, Latif E, et~al (2023) Multimodality of ai for education: Towards
  artificial general intelligence. arXiv preprint arXiv:231206037

\bibitem[{Lemke(2004)}]{lemke_literacies_2004}
Lemke JL (2004) The literacies of science. The Graduate Center, City University
  of New York

\bibitem[{Maier and Klotz(2022)}]{maier2022personalized}
Maier U, Klotz C (2022) Personalized feedback in digital learning environments:
  Classification framework and literature review. Computers and Education:
  Artificial Intelligence 3:100080

\bibitem[{Mason and Bruning(2001)}]{mason2001providing}
Mason BJ, Bruning R (2001) Providing feedback in computer-based instruction:
  What the research tells us. Retrieved February 15:2007

\bibitem[{Molenaar(2022)}]{molenaar2022towards}
Molenaar I (2022) Towards hybrid human-ai learning technologies. European
  Journal of Education 57(4):632--645

\bibitem[{Monteiro et~al(2021)Monteiro, Carvalho, and
  Santos}]{monteiro2021creating}
Monteiro V, Carvalho C, Santos NN (2021) Creating a supportive classroom
  environment through effective feedback: Effects on students’ school
  identification and behavioral engagement. In: Frontiers in Education,
  Frontiers Media SA, p 661736

\bibitem[{Moreno(2004)}]{moreno2004decreasing}
Moreno R (2004) Decreasing cognitive load for novice students: Effects of
  explanatory versus corrective feedback in discovery-based multimedia.
  Instructional science 32(1):99--113

\bibitem[{Mory(2004)}]{Mory2004}
Mory EH (2004) Feedback research revisited. In: Jonassen DH (ed) Handbook of
  Research on Educational Communications and Technology, 2nd edn. Lawrence
  Erlbaum Associates Publishers, p 745--783

\bibitem[{Nazaretsky et~al(2024)Nazaretsky, Mejia-Domenzain, Swamy, Frej, and
  K{\"a}ser}]{nazaretsky2024ai}
Nazaretsky T, Mejia-Domenzain P, Swamy V, et~al (2024) Ai or human? evaluating
  student feedback perceptions in higher education. In: European Conference on
  Technology Enhanced Learning, Springer, pp 284--298

\bibitem[{Nguyen et~al(2023)Nguyen, Stec, Hou, Di, and
  McLaren}]{nguyen2023evaluating}
Nguyen HA, Stec H, Hou X, et~al (2023) Evaluating chatgpt’s decimal skills
  and feedback generation in a digital learning game. In: European Conference
  on Technology Enhanced Learning, Springer, pp 278--293

\bibitem[{Ossenberg et~al(2019)Ossenberg, Henderson, and
  Mitchell}]{ossenberg2019attributes}
Ossenberg C, Henderson A, Mitchell M (2019) What attributes guide best practice
  for effective feedback? a scoping review. Advances in Health Sciences
  Education 24:383--401

\bibitem[{Sadler(1989)}]{sadler1989formative}
Sadler DR (1989) Formative assessment and the design of instructional systems.
  Instructional science 18(2):119--144

\bibitem[{Schultz(2019)}]{schultz2019effects}
Schultz M (2019) The effects of formative feedback on student learning in
  science education

\bibitem[{Se{\ss}ler et~al(2023)Se{\ss}ler, Xiang, Bogenrieder, and
  Kasneci}]{sessler2023peer}
Se{\ss}ler K, Xiang T, Bogenrieder L, et~al (2023) Peer: Empowering writing
  with large language models. In: European Conference on Technology Enhanced
  Learning, Springer, pp 755--761

\bibitem[{Se{\ss}ler et~al(2024{\natexlab{a}})Se{\ss}ler, F{\"u}rstenberg,
  B{\"u}hler, and Kasneci}]{sessler2024can}
Se{\ss}ler K, F{\"u}rstenberg M, B{\"u}hler B, et~al (2024{\natexlab{a}}) {Can
  AI grade your essays? A comparative analysis of large language models and
  teacher ratings in multidimensional essay scoring}. arXiv preprint
  arXiv:241116337

\bibitem[{Se{\ss}ler et~al(2024{\natexlab{b}})Se{\ss}ler, Rong,
  G{\"o}zl{\"u}kl{\"u}, and Kasneci}]{sessler2024benchmarking}
Se{\ss}ler K, Rong Y, G{\"o}zl{\"u}kl{\"u} E, et~al (2024{\natexlab{b}})
  Benchmarking large language models for math reasoning tasks. arXiv preprint
  arXiv:240810839

\bibitem[{Shute(2008)}]{shute2008focus}
Shute VJ (2008) Focus on formative feedback. Review of educational research
  78(1):153--189

\bibitem[{Tseng et~al(2024)Tseng, Yadav, Hou, Wu, Chou, Chen, Wu, Chen, Lin,
  Liao, and Koedinger}]{tseng2024effectiveness}
Tseng YJ, Yadav G, Hou X, et~al (2024) Activeai: The effectiveness of an
  interactive tutoring system in developing k-12 ai literacy. In:
  Ferreira~Mello R, Rummel N, Jivet I, et~al (eds) Technology Enhanced Learning
  for Inclusive and Equitable Quality Education. Springer Nature Switzerland,
  Cham, pp 452--467

\bibitem[{Vie et~al(2022)Vie, Rigaux, and Minn}]{vie2022privacy}
Vie JJ, Rigaux T, Minn S (2022) Privacy-preserving synthetic educational data
  generation. In: European Conference on Technology Enhanced Learning,
  Springer, pp 393--406

\bibitem[{Vladu{\v{s}}i{\'c} et~al(2016)Vladu{\v{s}}i{\'c}, Bucat, and
  O{\v{z}}i{\'c}}]{vladuvsic2016understanding}
Vladu{\v{s}}i{\'c} R, Bucat R, O{\v{z}}i{\'c} M (2016) Understanding of words
  and symbols by chemistry university students in croatia. Chemistry Education
  Research and Practice 17(3):474--488

\bibitem[{Wisniewski et~al(2020)Wisniewski, Zierer, and
  Hattie}]{wisniewski2020power}
Wisniewski B, Zierer K, Hattie J (2020) The power of feedback revisited: A
  meta-analysis of educational feedback research. Frontiers in psychology
  10:487662

\bibitem[{Wu et~al(2023)Wu, He, Liu, Liu, and Zhai}]{wu2023matching}
Wu X, He X, Liu T, et~al (2023) Matching exemplar as next sentence prediction
  (mensp): Zero-shot prompt learning for automatic scoring in science
  education. In: International conference on artificial intelligence in
  education, Springer, pp 401--413

\bibitem[{Zhai and Nehm(2023)}]{zhai2023ai}
Zhai X, Nehm RH (2023) Ai and formative assessment: The train has left the
  station. Journal of Research in Science Teaching 60(6):1390--1398

\bibitem[{Zhai et~al(2020)Zhai, Yin, Pellegrino, Haudek, and
  Shi}]{zhai2020applying}
Zhai X, Yin Y, Pellegrino JW, et~al (2020) Applying machine learning in science
  assessment: a systematic review. Studies in Science Education 56(1):111--151

\bibitem[{Zlabkova et~al(2024)Zlabkova, Petr, Stuchlikova, Rokos, and
  Hospesova}]{zlabkova2024development}
Zlabkova I, Petr J, Stuchlikova I, et~al (2024) Development of teachers'
  perspective on formative peer assessment. In: Developing Formative Assessment
  in STEM Classrooms. Routledge, p 105--125

\end{thebibliography}
